\documentclass[runningheads]{llncs}

\usepackage{eccv}

\usepackage{eccvabbrv}
\usepackage{multirow, framed}
\usepackage{graphicx}
\usepackage{wrapfig}
\usepackage{booktabs}
\usepackage{marvosym}
\usepackage[accsupp]{axessibility}  % Improves PDF readability for those with disabilities.
\usepackage[absolute,overlay]{textpos} % 使用 textpos 包

\usepackage[pagebackref,breaklinks,colorlinks,citecolor=eccvblue]{hyperref}
\usepackage{orcidlink}

\begin{document}

\title{Advancing Medical Radiograph Representation Learning: A Hybrid Pre-training Paradigm with Multilevel Semantic Granularity} 

\titlerunning{HybridMED}

\author{Hanqi Jiang$^*$\inst{1,2} \and
Xixuan Hao$^*$\inst{1} \and
Yuzhou Huang\inst{1} \and
Chong Ma\inst{3} \and
Jiaxun Zhang\inst{4} \and
Yi Pan\inst{2} \and
Ruimao Zhang\inst{1}\textsuperscript{\Letter} \\
$^*$ Equal Contribution.
}

\authorrunning{Jiang et al.}

\institute{The Chinese University of Hong Kong, Shenzhen \and
The University of Georgia \and
Northwestern Polytechnical University \and 
University of Illinois at Urbana-Champaign
}

\maketitle

\begin{abstract}
This paper introduces an innovative approach to Medical Vision-Language Pre-training (Med-VLP) area in the specialized context of radiograph representation learning. While conventional methods  frequently merge textual annotations into unified ``reports'', we acknowledge the intrinsic hierarchical relationship between the ``findings'' and ``impression'' section in radiograph datasets. To establish a targeted correspondence between images and texts, we propose a novel \texttt{HybridMED} framework to align global\-level visual representations with ``impression'' and token\-level visual representations with ``findings''. Moreover, our framework incorporates a generation decoder that employs two proxy tasks, responsible for generating the ``impression'' from (1) images, via a captioning branch, and (2) ``findings'', through a summarization branch. Additionally, knowledge distillation is leveraged to facilitate the training process. 
% Our method intertwines semantic hierarchies, generation, and distillation, aiming for enhanced performance in various medical downstream tasks. 
Experiments on the MIMIC-CXR dataset reveal that our summarization branch effectively distills knowledge to the captioning branch, enhancing model performance without significantly increasing parameter requirements due to the shared self-attention and feed-forward architecture. 
% We hope our framework could provid a meaningful contribution to Med-VLP methodologies within the medical imaging area.
\end{abstract}

%%%%%%%%%%%%%%%%%%%%%%%%%%%%%%%%%%%%%%%%%%%%%%%%%%%%%%%%%%%
%%%%%%%%%%%%%%%%%%%%%%%%%%%%%%%%%%%%%%%%%%%%%%%%%%%%%%%%%%%
% 1. Introduction
\section{Introduction}
\label{sec1}
The Vision-Language Pre-training (VLP) aims to effectively harness a massive amount of image-text pairs to comprehend a general multi-modal representation. A meticulously crafted multi-modal pre-training model can be effectively adapted to a wide range of downstream tasks, including, but not limited to, zero- and few-shot image classification, object detection, semantic segmentation, and visual question answering (VQA).
In the domain of radiograph representation learning, high-quality image-text datasets are notably scarce compared to those commonly available in the general computer vision community~\cite{jia2021scaling, zhai2022scaling}. This shortage arises primarily from the high costs associated with data acquisition, which frequently necessitate annotation by medical experts. Consequently, the effective pretrained models using existing open-source medical multi-modal datasets is of critical importance in advancing this field.

In recent years, the introduction of the MIMIC-CXR dataset~\cite{johnson2019mimic}, as a milestone, has significantly accelerated the progress of radiograph representation learning. This dataset includes chest X-ray images paired with medical reports, typically featuring a `findings' section detailing medical observations and an `impression' section summarizing key features of the radiograph.
Leveraging high-quality medical datasets, pioneering approaches such as ConVIRT~\cite{zhang2022contrastive}, GLoRIA~\cite{huang2021gloria}, and MGCA~\cite{wang2022multi} primarily rely on contrastive learning for pre-training, owing to the demonstrated efficacy~\cite{he2020momentum, chen2020improved, fan2021multiscale, chen2020simple, radford2021learning} in computer vision and multi-modal researches.
Meanwhile, certain studies have concentrated on the effective integration of contrastive learning and generative pre-text tasks~\cite{wang2021simvlm, yu2022coca}.
These studies indicate that pre-training can concurrently aid uni-modal tasks (e.g., fine-tuned classification), cross-modal tasks (e.g., zero-shot classification) and multi-modal tasks (e.g., VQA). 
% Inspired by these studies, our focus is on determining how to skillfully utilize the limited quantity of medical multi-modal datasets to collectively enhance various types of medical downstream tasks, through combining contrastive and generative paradigm.

\begin{wrapfigure}{r}{0.65\textwidth}
\vspace{-2.5em}
\centering
\includegraphics[scale=0.27]{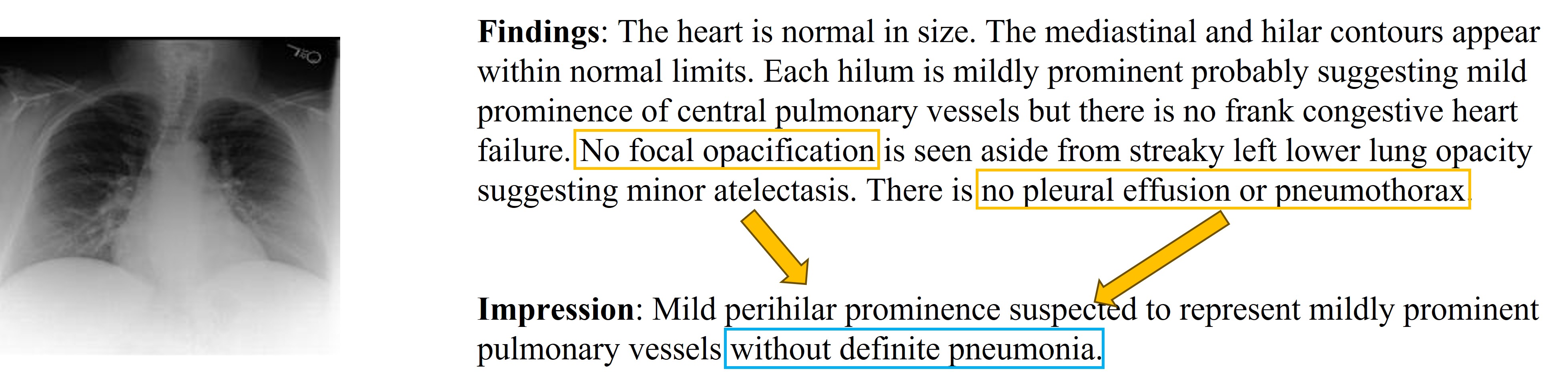}
\caption{An example of Semantic Hierarchy between radiograph ``findings'' and ``impression'' from MIMIC-CXR dataset.}
\label{fig:1-SemanticHierarchy}
\vspace{-2em}
\end{wrapfigure}
% The MIMIC-CXR dataset~\cite{johnson2019mimic} serves as a milestone in the field of radiograph representation learning, which contains chest X-ray images paired with corresponding medical reports. These reports typically include a ``findings'' section, which details medical observations, and an ``impression'' section, which summarizes the most remarkable features observed in the images. 
Though promising, the aforementioned contrastive learning frameworks in MedVLP typically suffer from two shortcomings: a) At the data level, they tend to directly concatenate ``findings'' and ``impression'', treating them equivalently. b) At the model level, they either simply align global tokens across both modalities~\cite{zhang2022contrastive}, or introduce a local contrastive branch that aligns regional visual features with word-level features~\cite{huang2021gloria, wang2022multi}.
Consequently, previous practices have ignored the fact that ``findings'' and ``impression'' represent two distinct semantic granularities with a hierarchical relationship that warrants further exploration.
As shown in Fig. \ref{fig:1-SemanticHierarchy}, the diagnosis of \texttt{no pneumonia} (high semantic level, providing a diagnosis for the overall disease) is derived from the combination of information from \texttt{no focal opacification} and \texttt{no pleural effusion or pneumothorax} (low semantic level, describing localized symptoms).
% a sample from the MIMIC-CXR dataset demonstrates the semantic hierarchy contained within the radiograph report. 
This hierarchy can provide valuable context for understanding the medical images, as it links specific observations with their broader significance. 
% The multi-modal data provided by the MIMIC-CXR dataset makes it a valuable resource for training models to understand and interpret medical images and reports.
% 由于image模态和多层级的text模态的存在，使得我们可以并行结合image captioning任务和text summarization任务，使其收敛于相同的结果——medical impression生成，从而从两个角度共同促进medical representation learning.

To this end, we believe that delving into this long-overlooked data semantic characteristic will significantly contribute to the multi-modal representation learning of Med-VLP.
To leverage the hierarchical attributes of radiograph reports with visual features, we propose \texttt{HybridMED}, which aims to explore the potential of a multi-level semantic granularity pre-training method in a joint contrastive-generative manner.
Our proposed \texttt{HybridMED} consists of three components. (1) The Contrastive Branch. 
We enforce a global-level alignment between the global image features and ``impression'' annotation. Additionally, we conduct a token-level alignment between multi-scale aggregated image features and ``findings'' annotation.
(2) The Multi-level Generative Branch. We distinguish between the ``findings'' and ``impression'' annotations to construct a multi-modal hybrid representation. 
\begin{wrapfigure}{r}{0.4\textwidth}
  \vspace{-1em}
  \centering
  \includegraphics[width=0.4\textwidth]{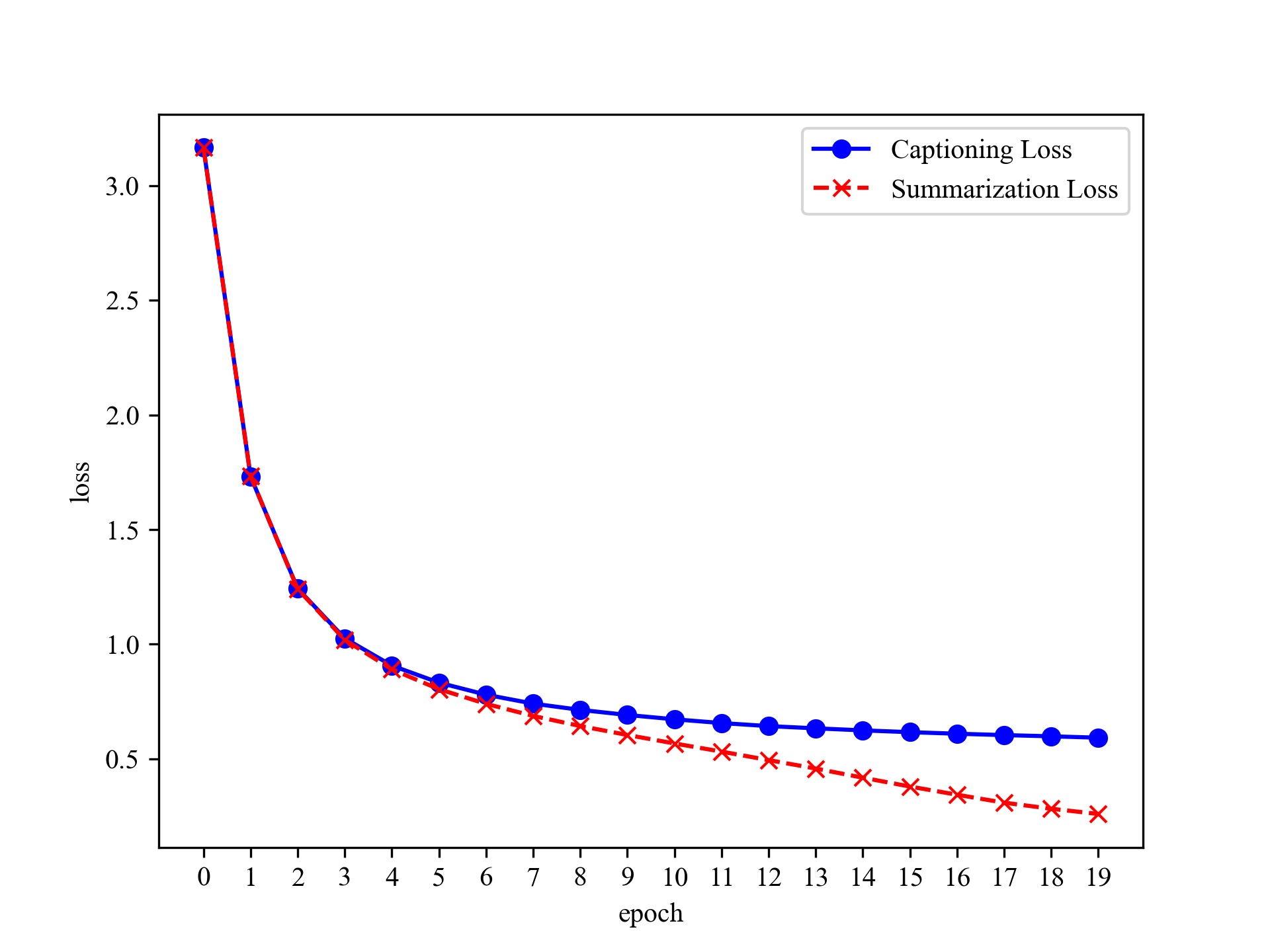} 
  \caption{Captioning Loss and Summarization Loss on MIMIC-CXR validation set, where better convergence in the summarization loss indicates that under equivalent generative objectives, captioning is a more challenging task.}
  \label{fig:2-CaptionSummaryLoss}
  \vspace{-3em}
\end{wrapfigure}
Our \texttt{HybridMED} framework incorporates two parallel generation branches. The first branch, a captioning module, generates the 'impression' from images, while the second, a summarization module, derives the 'impression' from the 'findings' section.
(3) The Collaborative Knowledge Distillation Branch. As evidenced by~\cite{hu2021word, hu2022graph, sotudeh2020attend} and Fig. \ref{fig:2-CaptionSummaryLoss},
% Several studies have shown that the 
the summarization task is typically easier than the captioning task in medical field~\cite{hu2021word, hu2022graph, sotudeh2020attend}. The difference lies in the fact that summarization is a uni-modal process, whereas deriving diagnostic conclusions about a patient's condition from medical imaging is a cross-modal task, requiring more rigorous reasoning informed by medical expertise.
As a result, a distillation mechanism is proposed to transfer knowledge from the summarization branch to aid the learning process of the captioning branch, which is utilized for multi-modal downstream tasks. This approach employs shared self-attention and feed-forward layers to enhance parameter efficiency.
% To substantiate this prior assumption, we execute an experiment on MIMIC-CXR dataset by MGCA baseline. This experiment incorporates two parallel branches for captioning and summarization, eschewing the use of knowledge distillation. The focus is solely on the generation of ``impression''. The generation losses from these two branches are illustrated in Fig. \ref{fig:2-CaptionSummaryLoss}. The diagram corroborates our hypothesis that the captioning task, when generating identical content, converges at a slower pace due to its inherent complexity.

In summary, we present a medical vision-language pre-training (Med-VLP) framework that incorporates multi-modal contrastive alignment and parallel generative streams with multi-level semantic hierarchies. To accomplish this goal, we effectively leverage the characteristics of medical data. By optimizing elaborate training objectives, our \texttt{HybridMED} is capable of efficiently executing a variety of downstream tasks, including cross-modal, uni-modal, and multi-modal types. Extensive experimental results demonstrate that our \texttt{HybridMED} can deliver highly satisfactory performance across a wide array of downstream tasks, thereby validating the model's superiority.

\section{Related Work}

\subsection{Vision-and-Language Pre-Training (VLP)}
%Self-supervised learning has garnered significant recognition in the realms of Computer Vision (e.g., MoCo~\cite{he2020momentum}, SimCLR~\cite{chen2020simple}, MAE~\cite{he2022masked}) and Natural Language Processing (e.g., BERT~\cite{devlin2018bert}, GPT1, 2 and 3~\cite{radford2018improving, radford2019language, brown2020language}). A well-structured pre-training framework can be effectively transferred to downstream tasks, thereby enhancing their performance. Concurrently, the advent of transformers has spurred the research community to extensively investigate the cross-attention mechanism for the amalgamation and interaction of different modalities. This has further propelled the advancement of multi-modal research. The paradigm of initial pre-training, followed by transfer to downstream tasks, remains a consistent approach in representation learning.

Self-supervised learning, recognized in Computer Vision (e.g., MoCo~\cite{he2020momentum}, SimCLR~\cite{chen2020simple}, MAE~\cite{he2022masked}) and Natural Language Processing (e.g., BERT~\cite{devlin2018bert}), benefits downstream tasks through effective pre-training frameworks. Concurrently, the advent of transformers has advanced multi-modal research, leveraging cross-attention mechanisms for the amalgamation and interaction of different modalities. The paradigm of initial pre-training, followed by transfer to downstream tasks, remains a consistent approach in representation learning.

%Early works on Vision-Language Pre-training (VLP) can be broadly categorized into single-stream and two-stream methodologies. Single-stream approaches~\cite{chen2020uniter, li2020oscar, li2019visualbert, su2019vl, li2020unicoder, kim2021vilt} employ a unified transformer architecture as a fusion module to seamlessly integrate two modalities. Conversely, two-stream methods~\cite{tan2019lxmert, lu2019vilbert, singh2022flava} initially utilize vision/language-specific encoders to extract features, and subsequently employ a fusion module to merge the two modalities. These distinct frameworks are optimized using a variety of pre-text tasks for pre-training, including masked-based language/image (MLM/MIM) and image-text matching (ITM).

Early works on Vision-Language Pre-training (VLP) can be broadly categorized into single-stream and two-stream methodologies. Single-stream approaches~\cite{chen2020uniter, li2020oscar, li2019visualbert, su2019vl, li2020unicoder, kim2021vilt} employ a unified transformer architecture as a fusion module, while two-stream methods~\cite{tan2019lxmert, lu2019vilbert, singh2022flava} initially utilize vision/language-specific encoders to extract features, and subsequently employ a fusion module to merge the two modalities. These distinct frameworks are optimized using a variety of pre-text tasks for pre-training, including masked-based language/image (MLM/MIM) and image-text matching (ITM).

The groundbreaking work of CLIP~\cite{radford2021learning} exemplifies the potent capabilities of the contrastive-based dual-encoder framework in cross-modal downstream tasks, such as zero-shot classification and cross-modal retrieval. Numerous related variants that delve into more granular multi-modal representations have been explored, including DeCLIP~\cite{li2021supervision}, FILIP~\cite{yao2021filip}, SLIP~\cite{mu2022slip}, etc.

Furthermore, some research ~\cite{li2022blip, wang2021simvlm, yu2022coca, alayrac2022flamingo} has started to explore the potential of unifying VLP by merging the dual-encoders with a fusion module. Specifically, these studies enhance the framework by optimizing it with contrastive loss and Language Model (LM) loss, which can be generally transferred to more types of downstream tasks. This includes not only uni-modal or cross-modal downstream tasks like supervised learning classification, detection, segmentation, and zero-shot classification, but also multi-modal tasks like Visual Question Answering (VQA) that require vision-language interaction. This paper presents the model architecture characterized by dual-encoders, comprising an image encoder and a uni-modal text decoder, as well as fusion modules. These fusion modules are represented by a captioning branch, which is further assisted by a summarization branch.

\subsection{Medical Vision-and-Language Pre-Training (Med-VLP)}
Med-VLP is a specific division of VLP in the medical domain, aims to exploit large-scale multi-modal medical datasets to jointly represent both radiographs and reports. 
Early methods employ dual-encoders to globally align these two modalities~\cite{zhang2022contrastive}, or to extract word-patch features and conduct additional alignment in a local manner~\cite{huang2021gloria}. Subsequent improvements related to dual-encoders, such as MGCA~\cite{wang2022multi}, introduces a triplet alignment encompassing pathological region-level, instance-level, and disease-level. BioViL~\cite{boecking2022making} initially emphasizes the effectiveness of BERT~\cite{devlin2018bert} trained on dedicated medical texts, as opposed to a common medical text encoder, and the pre-trained BERT is further aligned with images to achieve superior performance.

In addition, innovative methods focusing on fusion modules have also been developed. For instance, M3AE~\cite{chen2022multi} introduces a multi-modal fusion encoder under MIM and MLM training objectives, while ARL~\cite{chen2022align} subsequently integrates an external medical knowledge graph, namely UMLS~\cite{bodenreider2004unified}, in the pre-training stage to enhance its representation ability. PTUnifier~\cite{chen2023towards} incorporates both dual-encoders and the fusion module, seeks for the extensionality and generalization of Med-VLP. These works have yielded promising results across a wide range of downstream tasks.

While aforementioned methods simply utilize the whole medical reports for representation learning, ours, on the other hand, exploring to encapsulate multi-level semantic granularity, tailored to the unique characteristics of medical data. Our overarching aim is to unify a diverse range of medical downstream tasks.

%%%%%%%%%%%%%%%%%%%%%%%%%%%%%%%%%%%%%%%%%%%%%%%%%%%%%%%%%%%
%%%%%%%%%%%%%%%%%%%%%%%%%%%%%%%%%%%%%%%%%%%%%%%%%%%%%%%%%%%
% 3. Methodology
\section{Methodology}
\label{gen_inst}
\begin{figure*}[h!]
\centering
\includegraphics[scale=0.24]{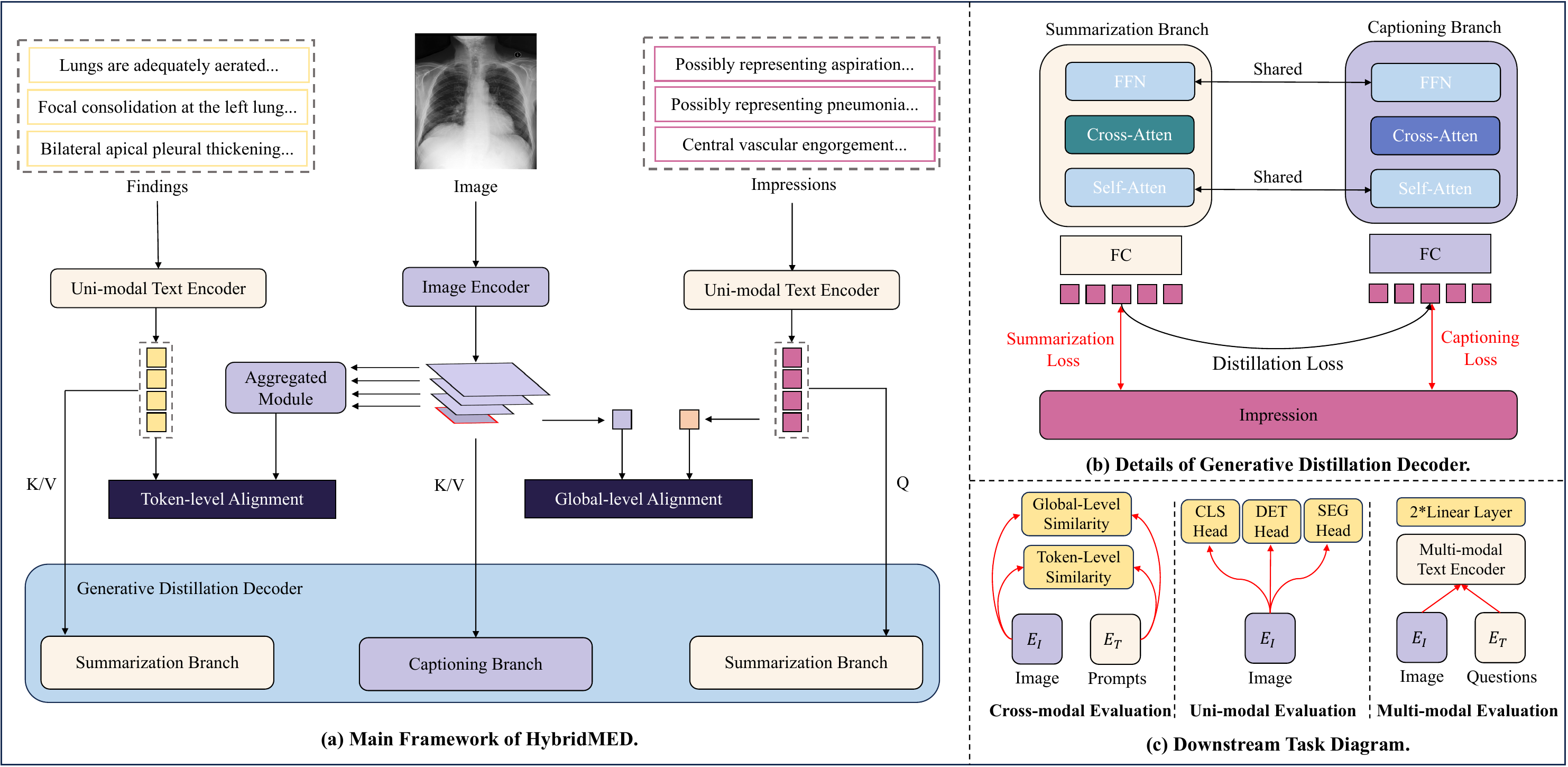}
\caption{The \texttt{HybridMED} framework is presented in two parts. (a) introduces the overall framework, which encompasses multi-modal alignment across multi-level semantic hierarchies and parallel generative distillation decoders. (b) delves into the specifics of the two parallel generative decoders. The self-attention layers and feed-forward layers in these two branches share weights, while the cross-attention layers differ, conditioned on different modalities. Furthermore, the summarization branch distills its outputs to facilitate the operations of the captioning branch. (c) describes multiple downstream tasks diagram.}
\label{fig:3-HybridMED}
\vspace{-2em}
\end{figure*}

% 需要修改章节引用
In this paper, we present \texttt{HybridMED}, a framework specifically designed for hybrid medical multi-modal representation learning with multi-level semantic granularity. The framework is shown in Fig. \ref{fig:3-HybridMED}. In Sec. \ref{sec3.1}, we firstly introduce the global- and token-level contrastive alignment modules. Subsequently, in Sec. \ref{sec3.2}, we discuss the construction of two parallel generative branches, utilizing knowledge distilled from the summarization branch to enhance the captioning branch. Finally, in Sec. \ref{sec3.3}, we summarize the comprehensive training objectives of our \texttt{HybridMED} framework.

% \begin{figure*}[h!]
% \centering
% \includegraphics{img/3-HybridMED.jpg}
% \caption{The HybridMED framework is presented in two parts. (a) introduces the overall framework, which encompasses multi-modal alignment across multi-level semantic hierarchies and parallel generative distillation decoders. (b) delves into the specifics of the two parallel generative decoders. The self-attention layers and feed-forward layers in these two branches share weights, while the cross-attention layers differ, conditioned on different modalities. Furthermore, the summarization branch distills its outputs to facilitate the operations of the captioning branch.}
% \label{fig:3-HybridMED}
% \end{figure*}

% 3.1. Multi-Modal Alignment across Semantic Hierarchies
\subsection{Multi-Modal Alignment across Semantic Hierarchies}
\label{sec3.1}
% \textbf{Global-level Alignment}
% \textbf{Token-level Alignment}
% $W=\{(x_{i}),(y_{i}), (z_{i})\}$
Given a set of image-report pairs $W=\{x, y, z\}$, where $x$ represents an image, $y$ and $z$ denote the corresponding ``impression'' and “findings”, respectively, our objective is to align paired image-report in latent spaces, bringing similar ones closer while pushing dissimilar ones farther apart. Given the unique characteristics of medical reports, the term ``impression'' denotes a diagnosis formulated by physicians based on comprehensive descriptions of symptoms, serving as the textual global guidance. Conversely, ``findings'' encompass more semantic meanings, thereby encapsulating disease-level information. We partition the reports into ``findings'' and ``impression'', and subsequently propose the global-token contrastive alignment architecture with multi-level semantic visual representation.
% $x_{ti\_F}$, $x_{ti\_I}$

To establish a global-level semantic correspondence between images and their associated ``impression'', we employ a joint optimization approach for the image encoder and the uni-modal text decoder. This is achieved by contrasting the paired data against other data within the sampled batch:

\begin{equation}
\mathcal{L}_{CG}^{x|y} = -\sum_i^N{{\rm log} \frac{((x_{gi}^{\top}y_{gi})/\tau)}{(\sum_{j=0}^{N}{{\rm exp}(x_{gi}^{\top}y_{gj})}/\tau)}}
\end{equation}

\begin{equation}
\mathcal{L}_{CG}^{y|x} = -\sum_i^N{{\rm log} \frac{((y_{gi}^{\top}x_{gi})/\tau)}{\sum_{j=0}^{N}{{\rm exp}(y_{gi}^{\top}x_{gj})}/\tau)}}
\end{equation}

\begin{equation}
\mathcal{L}_{CG} = \frac{1}{N}(\mathcal{L}_{CG}^{x|y} + \mathcal{L}_{CG}^{y|x})
\end{equation}

where $x_{gi}$ and $y_{gj}$ are the normalized embeddings of the average pooling feature in the i-th image and that of the class token in the j-th “impression”, respectively. Besides, $N$ is the batch size, and $\tau$ is the temperature to scale the logits.

Furthermore, to construct alignment between images and their associated ``findings'', we firstly consolidate the multi-scale image features by the aggregate modules, which involves the Feature Pyramid Pooling (FPN) modules~\cite{lin2017feature} and two convolutional layers. We denote $\{(x_{si_{1}}), ... , (x_{si_{m}})\}$ as the varying scale of visual features and $x'_{i}$ indicates aggregated image features of the i-th image:

\begin{equation}
x'_{i} = 2 \ast Conv(FPN(x_{si_{1}} ... x_{si_{m}}))
\end{equation}

Drawing inspiration from FILIP~\cite{yao2021filip}, we further leverage the fine-grained contrastive expressiveness based on the mutual average token-wise maximum similarity between the two modalities. Specifically, we initially calculate the similarity of each visual token with all textual tokens, utilizing the highest value to compute the average similarity of all image tokens to textual ones. Notably, this process is bi-directional, implying that the same procedure will be executed by interchanging image and text tokens:

\begin{equation}
sim(x'_{i}, z_{i}) = \frac{1}{n_1}(\sum_{k=1}^{n_1}{x'_{ki}{}^{\top}\mathop{{\rm argmax}}\limits_{k \in [0,n_2)}(z_{ki})})
\end{equation}

\begin{equation}
sim(z_{i}, x'_{i}) = \frac{1}{n_2}(\sum_{k=1}^{n_2}{z_{ki}^{\top}\mathop{{\rm argmax}}\limits_{k \in [0,n_1)}(x'_{ki})})
\end{equation}

where $\{(z_{1i}), ... , (z_{ki})\}$ denote the token-level features from ``findings'', and $n_1$ and $n_2$ are denoted as the number of tokens of the i-th aggregated image features and j-th ``findings'', and the fine-grained token-level representation could be finally formulated as:

\begin{equation}
\mathcal{L}_{CL}^{x|z} = -\sum_i^N{{\rm log}\frac{(sim(x'_{i}, z_{i}))/\tau)}{(\sum_{j=0}^N{{\rm exp}(sim(x'_{i}, z_{j}))}/\tau)}}
% \sum_{j=0}^{N}{{\rm exp}((x_{gi}^{\top}y_{gj})/\tau)}
\end{equation}

\begin{equation}
\mathcal{L}_{CL}^{z|x} = -\sum_i^N{{\rm log}\frac{(sim(z_{i}, x'_{i}))/\tau)}{(\sum_{j=0}^N{{\rm exp}(sim(z_{i}, x'_{j}))}/\tau)}}
\end{equation}

\begin{equation}
\mathcal{L}_{CL} = \frac{1}{N}(\mathcal{L}_{CL}^{x|z} + \mathcal{L}_{CL}^{z|x})
\end{equation}

\subsection{Generative Distillation Decoder}
\label{sec3.2}
The multi-modal text decoder is designed to address multi-modal understanding, necessitating interaction between visual and textual modalities. Instead of merely constructing it by generating the whole reports, we introduce two parallel generative branches, with knowledge distilled from the summarization branch to assist the captioning branch, drawing on prior experiences from medical NLP researches.

Specifically, we initially construct the uni-modal summarization branch (abbreviated to summarization branch) by generating ``impression'' conditioned on ``findings'', utilizing the cross-attention mechanism~\cite{vaswani2017attention}. In this context, ``impression'' functions as the query, while ``findings'' serves as the key and value. Additionally, the multi-modal captioning branch (abbreviated to captioning branch) is established by generating identical ``impression'' conditioned on image features.

Therefore, for these two branches, we both train the maximum log-likelihood objective. This approach captions the ``impression'' through a teacher-forcing strategy. Consequently, the objectives for summarization and captioning processes can be independently formulated as follows:

\begin{equation}
\mathcal{L}_{Sum}(\theta_{1}) = -\sum_{t=1}^T{{\rm log}p_{\theta_{1}}(y_{ti}|y_{(0:t-1)i}, (z_{1i} ... z_{ki}))}
\end{equation}

\begin{equation}
\mathcal{L}_{Cap}(\theta_{2}) = -\sum_{t=1}^T{{\rm log}p_{\theta_{2}}(y_{ti}|y_{(0:t-1)i}, x_{gi})}
\end{equation}

where $\{(z_{1i}), ... , (z_{ki})\}$ and $x_{gi}$ are still the previous definitions, $T$ denotes the token number of ``impression'', and $\theta_{1}$ and $\theta_{2}$ indicate the model of summarization and captioning branches, respectively. Notice that for parameter efficiency, the weights associated with self-attention and feed-forward layers are shared across the two branches.

The summarization branch has the potential to further distill its outputs to aid in the generation of the ``impression'' within the captioning branch, where the lateral is served as the central architecture for multi-modal understanding in downstream tasks. We employ the Kullback-Leibler (KL) divergence for this purpose, and the distillation objective can be articulated as follows:

\begin{equation}
P_{Sum} = p_{\theta_{1}}(y_{ti}|y_{(0:t-1)i}, (z_{1i} ... z_{ki}))
\end{equation}

\begin{equation}
P_{Cap} = p_{\theta_{2}}(y_{ti}|y_{(0:t-1)i}, x_{gi})
\end{equation}

\begin{equation}
\mathcal{L}_{Dis} = \sum_{t=1}^T{P_{Sum_{t}}}{{\rm log}}\frac{{P}_{Sum_{t}}} {P_{Cap_{t}}}
\end{equation}

% 3.3. HybridMED Pre-Training Objective
\subsection{HybridMED Pre-Training Objective}
\label{sec3.3}
In the end, we construct our \texttt{HybridMED} by integrating the multi-level semantic alignment module with the parallel generative distillation decoder. The comprehensive pre-training objective can be expressed as follows:

\begin{equation}
\mathcal{L} = \lambda_{CG}\mathcal{L}_{CG} 
+ \lambda_{CL}\mathcal{L}_{CL} 
+ \lambda_{Sum}\mathcal{L}_{Sum} 
+ \lambda_{Cap}\mathcal{L}_{Cap}
+ \lambda_{Dis}\mathcal{L}_{Dis}
\end{equation}

%%%%%%%%%%%%%%%%%%%%%%%%%%%%%%%%%%%%%%%%%%%%%%%%%%%%%%%%%%%
%%%%%%%%%%%%%%%%%%%%%%%%%%%%%%%%%%%%%%%%%%%%%%%%%%%%%%%%%%%
% 4. Experiments
\section{Experiments}
\label{headings}

% 4.1. Experimental Settings
\subsection{Experimental Settings}
\textbf{Training Process.} The training procedure for our \texttt{HybridMED} is divided into two primary stages. In the first stage, we only train the summarization branch, excluding the contrastive objectives and the captioning branch. In the second stage, we establish the global-token contrastive alignment. This is done in conjunction with the generative distillation decoder, where we freeze the summarization branch since it acts as a teacher to assist the student captioning branch.
% the self-attention and FFN layers are also frozen

\textbf{Network Architecture.} Referring to the settings of MGCA~\cite{wang2022multi}, the image encoder is implemented with ResNet50~\cite{he2016deep}, and we aggregate the multi-scale image features using the Feature Pyramid Pooling (FPN) network~\cite{lin2017feature}, for which allows us to extract features with resolutions of 8x8, 16x16, 32x32 and 64x64. Subsequently, there are two 3x3 convolutional neural layers to downsample the features for the token-level alignment. For the textual backbone, we initialize it using a 12-layer BioClinicalBERT~\cite{alsentzer2019publicly}. We divide the first 6-layer transformers into the uni-modal text decoder, and the remaining 6-layer transformers are used as the two decoders. We further insert 6-layer cross attention transformers into the summarization and captioning branches under different conditions, respectively.

\textbf{Implementation Details.} All the experiments are conducted on NVIDIA A100 GPUs, and both stages are trained for 50 epochs with early stopping. The batch size is set to 48. During the first stage of training, only the summarization loss is involved, where the AdamW optimizer~\cite{loshchilov2017decoupled} is used, and the learning rate and weight decay parameters are set to 2e-5 and 0.05, respectively.
In the second stage, we also adopt the AdamW optimizer. The values of the learning rate, weight decay, and warm-up epoch are set to 2e-5, 0, and 20, respectively. In this phase, the loss function is composed of all five parts, as mentioned in Section 3.3, and all the loss weights are set to 1.

% 4.2. Pretraining Dataset
\subsection{Pretraining Dataset}
MIMIC-CXR~\cite{johnson2019mimic} is one of the largest open-source medical multi-modal dataset available for radiograph representation learning, compiled from routine clinical practices. This dataset comprises approximately 232k chest X-ray images, encompassing both frontal and lateral views, along with 367k reports. These reports primarily consist of ``findings'' and ``impression''. During the pre-processing stage, we initially exclude all lateral view scans and eliminate cases with empty ``findings'' and ``impression''. This results in a refined dataset of approximately 135k image-report pairs.

% \begin{figure*}[h!]
% \centering
% \includegraphics[scale=0.4]{img/4-Downstream.jpg}
% \caption{Downstream tasks diagram.}
% \label{fig:4-Downstream}
% \end{figure*}

% 4.3. Downstream Tasks Datasets
\subsection{Downstream Tasks Datasets}
We conduct extensive downstream tasks to evaluate our \texttt{HybridMED}, and we first introduce datasets used for different tasks.
(1) \textbf{RSNA Pneumonia}~\cite{shih2019augmenting} is a versatile dataset, which involves about 29.7k frontal view chest radiographs. This dataset is binary (i.e., normal or pneumothorax positive) that can be used for zero- or few-shot image classification, object detection and semantic segmentation. The strategies for data splitting vary across these tasks and will be individually detailed in the following parts.
(2) \textbf{CheXpert}~\cite{irvin2019chexpert} comprises 191,229 frontal chest radiographs, which can be utilized for five distinct binary classifications, specifically: atelectasis, cardiomegaly, consolidation, edema, and pleural effusion. We employ the expert-labeled validation set as test data, and randomly select 5,000 samples from the training set for validation purposes.
% (3) \textbf{Object-CXR}~\cite{healthcare2020object} encompasses 9,000 frontal-view chest X-rays, each annotated with bounding boxes for foreign object detection. We conduct evaluations using the original test set, which includes 1,600 samples. Additionally, we randomly divide the original training set in an 8:2 ratio to create a new training set and validation set, consisting of 6,400 and 1,600 samples respectively.
% (4) \textbf{SIIM Pneumothorax}~\cite{siim-acr-pneumothorax-segmentation} is utilized to evaluate the performance of semantic segmentation. It consists of 12,047 chest radiographs, each accompanied by manually annotated segmentation masks for pneumothorax. The dataset is also divided into training, validation, and testing subsets at a ratio of 7:3:3.
(3) \textbf{VQA-RAD}~\cite{lau2018dataset} and (4) \textbf{Med-VQA2019}~\cite{ben2019vqa} are two datasets that consist of 3,515 and 15,292 image-question pairs respectively. Although these datasets encompass multi-organ components, our primary focus is on chest studies, in line with our radiograph representation learning. In addition, we adhere to the splitting and processing settings as outlined by PTUnifier~\cite{chen2023towards}.
% \footnote{https://www.kaggle.com/c/siim-acr-pneumothorax-segmentation} 

% 4.4. Results of Downstream Tasks
\subsection{Results of Downstream Tasks}
Our \texttt{HybridMED} is designed to uniformly address a variety of downstream tasks, including cross-modal, uni-modal, and multi-modal tasks. 

% 4.4.1
\subsubsection{Cross-modal Evaluation}
Cross-modal evaluation involves assessing the interactions between different types of data, particularly between visual and textual modalities, to enhance the overall performance of the model. We primarily evaluates the trained model on zero-shot classification tasks.

\textbf{Zero-Shot Classification.} Cross-modal evaluation primarily involves zero-shot classification, necessitating a robust alignment between visual and textual modalities. This task is conducted on two datasets: the RSNA Pneumonia dataset (binary classification) and the CheXpert 5x200 dataset (five categories). For the RSNA Pneumonia dataset, we adopt the settings outlined in BioViL~\cite{boecking2022making} to construct four positive and negative prompts respectively, such as ``Findings suggesting pneumonia'' and ``No evidence of pneumonia''. The accuracy of this approach is evaluated on its test set split, comprising 8006 samples. For CheXpert dataset, we additionally follow GLoRIA~\cite{huang2021gloria} to extract a small-scale subset, CheXpert 5x200, which includes five distinct diseases: Atelectasis, Cardiomegaly, Edema, Pleural, and Effusion. Each disease category contains 200 exclusively positive images, accompanied by both positive and negative prompts.

\begin{wraptable}{r}{0.5\textwidth}
\vspace{-2em}
\centering
\resizebox{0.5\textwidth}{!}{
\begin{tabular}{cccc} 
    \toprule
    Method & Pretrain Dataset & RSNA & CheXpert 5×200 \\
    \midrule
    CLIP \cite{radford2021learning} & ImageNet & 0.250 & 0.201 \\
    ConVIRT \cite{zhang2022contrastive} & MIMIC-CXR & 0.719 & 0.213 \\
    GLoRIA-MIMIC \cite{huang2021gloria} & MIMIC-CXR & 0.730 & 0.248 \\
    PRIOR \cite{cheng2023prior} & MIMIC-CXR & 0.768 & 0.349 \\
    BioViL \cite{boecking2022making} & MIMIC-CXR & 0.732 & 0.354 \\
    MGCA \cite{wang2022multi} & MIMIC-CXR & \underline{0.781} & \underline{0.422} \\
    \midrule    
    Ours & MIMIC-CXR & \textbf{0.800} & \textbf{0.448} \\
    \bottomrule
\end{tabular}
}
\caption{Zero-Shot Classification results on RSNA Pneumonia and CheXpert 5×200 datasets (Acc). \textbf{Bold} denotes the best result and \underline{Underline} denotes the second-best result.}
\label{table:ZeroShotClassification}
 \vspace{-2.5em}
\end{wraptable}

We compute the image-text similarities on both global-level and token-level representations, and find out the category with the highest average similarity.
The results derived from these two datasets are presented in Table \ref{table:ZeroShotClassification}. Upon comparison with other methodologies, it is evident that our \texttt{HybridMED} achieves state-of-the-art results on both datasets in zero-shot classification. This underscores the effectiveness of multi-modal alignment in representing multi-level semantic granularity.
\subsubsection{Uni-modal Evaluation}
Uni-modal evaluation tasks include fine-tuned image classification, object detection, and semantic segmentation. In these tasks, the image encoder is always frozen, and the task-specific heads are optimized. In addition, we evaluate performances across varying proportions of training data, specifically 1\%, 10\%, and 100\%. Apart from the Object-CXR dataset for object detection, we only carry out transfer learning experiments using 10\% and 100\% of the training data. All the configurations for these tasks adhere to MGCA.

\textbf{Image Classification.} We conduct image classification on the RSNA Pneumonia and CheXpert datasets. In this process, we optimize a linear classification head that has been randomly initialized, and subsequently report the Area Under Curve (AUC) for both datasets. The corresponding results are presented in Table \ref{table:ImageClassification}. When compared to existing methods, our \texttt{HybridMED} model exhibits superior performance on both RSNA Pneumonia dataset and CheXpert dataset.
% Table-2: image classification
\begin{table*}
\begin{center}
\resizebox{0.8\textwidth}{!}{
%表格中的数据居中，c的个数为表格的列数
\begin{tabular}{cccccccc}
	\toprule
	\multirow{2}{*}{Method} & \multirow{2}{*}{Pretrain Dataset} & \multicolumn{3}{c}{RSNA (AUC)} & \multicolumn{3}{c}{CheXpert (AUC)} \\
 &  & 1\% & 10\% & 100\% & 1\% & 10\% & 100\% \\
 	\midrule
        CLIP \cite{radford2021learning} & ImageNet  & 0.749 & 0.745 & 0.763 & 0.744 & 0.797 & 0.814 \\
        VSE++  & CheXpert & 0.503 & 0.512 & 0.524 & 0.494 & 0.572 & 0.679 \\
        GLoRIA \cite{huang2021gloria} & CheXpert & 0.866 & 0.878 & 0.881 & 0.836 & 0.874 & 0.883 \\
  	ConVIRT \cite{zhang2022contrastive} & MIMIC-CXR & 0.774 & 0.801 & 0.813 & 0.859 & 0.868 & 0.873 \\
	GLoRIA-MIMIC \cite{huang2021gloria} & MIMIC-CXR & 0.865 & \underline{0.890} & 0.897 & 0.862 & 0.871 & 0.870 \\
        LoVT \cite{muller2022joint} & MIMIC-CXR & 0.855 & 0.865 & 0.893 & 0.851 & 0.881 & 0.883 \\
 	BioViL \cite{boecking2022making} & MIMIC-CXR & \underline{0.881} & 0.884 & 0.891 & 0.808 & 0.875 & \underline{0.884} \\
  	MGCA \cite{wang2022multi} & MIMIC-CXR & 0.858 & 0.877 & \underline{0.893} & 0.856 & 0.877 & 0.883 \\
        PRIOR \cite{cheng2023prior} & MIMIC-CXR & 0.857 & 0.871 & 0.892 & \underline{0.862} & \underline{0.883} & \underline{0.886} \\
        
        \midrule	
 	Ours & MIMIC-CXR & \textbf{0.884} & \textbf{0.892} & \textbf{0.902} & \textbf{0.872} & \textbf{0.888} & \textbf{0.893} \\
	\bottomrule
\end{tabular}
}
\end{center}
\caption{Image classification results on RSNA Pneumonia and CheXpert datasets with 1\%; 10\%; 100\% training data. \textbf{Bold} denotes the best result and \underline{Underline} denotes the second-best result.}
\label{table:ImageClassification}
\vspace{-20pt}
\end{table*}

% \vspace{-20pt}
\textbf{Object Detection.} Object detection task is conducted on the RSNA Pneumonia dataset. The aim was to predict the bounding boxes of pneumonia. The training set for RSNA Pneumonia is randomly split into 16k for training, 5.3k for validation, and 5.3k for testing. The YOLOv3 architecture ~\cite{redmon2018yolov3} is used for this task. We leverage YOLOv3 architecture and evaluate the performances on Mean Average Precisions (mAP), with Intersection Over Union (IOU) thresholds 0.4, 0.45, 0.5, 0.55, 0.6, 0.65, 0.7, 0.75. According to the results presented in Table \ref{tab:ObjectDetection}, ours achieves the best results under 1\% and 100\% training data fine-tuning on the RSNA Pneumonia dataset. 
% Table-3: object detection
% \begin{table}
% \begin{center}
% \resizebox{0.6\textwidth}{!}{
% %表格中的数据居中，c的个数为表格的列数
% \begin{tabular}{ccccc}
% 	\toprule
% 	\multirow{2}{*}{Method}  & \multirow{2}{*}{Pretrain Dataset} & \multicolumn{3}{c}{RSNA} &  \\
% &  & 1\% & 10\% & 100\% \\
%  	\midrule
%         CLIP \cite{radford2021learning} & ImageNet & - & 0.079 & 0.216 \\
%   	ConVIRT \cite{zhang2022contrastive}  & MIMIC-CXR & 0.082 & 0.156 & 0.179 \\
%         GLoRIA-MIMIC \cite{huang2021gloria} & MIMIC-CXR & 0.116 & 0.161 & 0.248 \\	
%         LoVT \cite{muller2022joint} & MIMIC-CXR & \underline{0.130} & 0.175 & 0.218 \\	
%         BioViL \cite{boecking2022making} & MIMIC-CXR  & 0.123 & 0.168 & 0.229  \\	
%         MGCA \cite{wang2022multi} & MIMIC-CXR & 0.129 & 0.168 & \underline{0.249}  \\
%         PRIOR\cite{cheng2023prior} & MIMIC-CXR & - & \textbf{0.196} & 0.222 \\
        
%   	\midrule
%         Ours & MIMIC-CXR & \textbf{0.166} & \underline{0.188} & \textbf{0.256}  \\        
% 	\bottomrule
% \end{tabular}
% }
% \end{center}
% \caption{Object Detection results on RSNA Pneumonia with 1\%; 10\%; 100\% training data, and Object-CXR datasets with 10\%; 100\% training data. \textbf{Bold} denotes the best result and \underline{Underline} denotes the second-best result.}
% \label{tab:ObjectDetection}
% \vspace{-20pt}
% \end{table}

% \vspace{-20pt}
\textbf{Semantic Segmentation.} Semantic Segmentation task is performed on the RSNA Pneumonia dataset, with the objective of predicting the segmentation masks for pneumonia. For the RSNA Pneumonia dataset, we maintain the same splitting scheme as used in object detection. The U-Net~\cite{ronneberger2015u} framework is employed for this task, and the Dice scores are reported as the evaluation metrics. As illustrated in Table \ref{tab:SemanticSegmentation}, our \texttt{HybridMED} model achieves the highest performance under all splitting plans for the RSNA Pneumonia dataset. This demonstrates the superior token-level representation of our model for pixel-level prediction tasks.
% Table-4: semantic segmentation
% \begin{table}
% \begin{center}
% \resizebox{0.6\textwidth}{!}{
% %表格中的数据居中，c的个数为表格的列数
% \begin{tabular}{ccccc}
% 	\toprule
% 	\multirow{2}{*}{Method} & \multirow{2}{*}{Pretrain Dataset} & \multicolumn{3}{c}{RSNA}  \\
%  & & 1\% & 10\% & 100\%  \\
%  	\midrule	
%         CLIP \cite{radford2021learning} & ImageNet & 0.348 & 0.399 & 0.640 \\
%   	ConVIRT \cite{zhang2022contrastive} & MIMIC-CXR & 0.550 & 0.674 & 0.675 \\
%         GLoRIA-MIMIC \cite{huang2021gloria} & MIMIC-CXR & 0.603 & \underline{0.687}	& 0.683 \\
%         LoVT \cite{muller2022joint} & MIMIC-CXR & 0.624 & 0.681 & 0.696  \\
%         BioViL \cite{boecking2022making} & MIMIC-CXR & 0.597 & 0.676 & 0.679  \\
% 	MGCA \cite{wang2022multi} & MIMIC-CXR & \underline{0.630} & 0.683 &	\underline{0.698}  \\
        
%   	\midrule	
%         Ours & MIMIC-CXR & \textbf{0.686} & \textbf{0.696} & \textbf{0.726} \\
% 	\bottomrule
% \end{tabular}
% }
% \end{center}
% \caption{Semantic Segmentation results on RSNA Pneumonia and SIIM Pneumothorax datasets with 1\%; 10\%; 100\% training data. \textbf{Bold} denotes the best result and \underline{Underline} denotes the second-best result.}
% \label{tab:SemanticSegmentation}
% \vspace{-20pt}
% \end{table}

\begin{table}[ht]
\centering
\begin{subtable}{0.48\textwidth}
\centering
\resizebox{\textwidth}{!}{
\begin{tabular}{ccccc}
    \toprule
    \multirow{2}{*}{Method}  & \multirow{2}{*}{Pretrain Dataset} & \multicolumn{3}{c}{RSNA} \\
    &  & 1\% & 10\% & 100\% \\
    \midrule
        CLIP \cite{radford2021learning} & ImageNet & - & 0.079 & 0.216 \\
        ConVIRT \cite{zhang2022contrastive}  & MIMIC-CXR & 0.082 & 0.156 & 0.179 \\
        GLoRIA-MIMIC \cite{huang2021gloria} & MIMIC-CXR & 0.116 & 0.161 & 0.248 \\  
        LoVT \cite{muller2022joint} & MIMIC-CXR & \underline{0.130} & 0.175 & 0.218 \\  
        BioViL \cite{boecking2022making} & MIMIC-CXR  & 0.123 & 0.168 & 0.229  \\   
        MGCA \cite{wang2022multi} & MIMIC-CXR & 0.129 & 0.168 & \underline{0.249}  \\
        PRIOR\cite{cheng2023prior} & MIMIC-CXR & - & \textbf{0.196} & 0.222 \\
    \midrule
        Ours & MIMIC-CXR & \textbf{0.166} & \underline{0.188} & \textbf{0.256}  \\        
    \bottomrule
    \addlinespace[5ex] % 增加空白行，以便表格高度一致
\end{tabular}
}
\caption{Object Detection results on RSNA Pneumonia with 1\%; 10\%; 100\% training data, and Object-CXR datasets with 10\%; 100\% training data. \textbf{Bold} denotes the best result and \underline{Underline} denotes the second-best result.}
\label{tab:ObjectDetection}
\end{subtable}%
\hfill
\begin{subtable}{0.48\textwidth}
\centering
\vspace*{-10ex} % 调整顶部对齐
\resizebox{\textwidth}{!}{
\begin{tabular}{ccccc}
    \toprule
    \multirow{2}{*}{Method} & \multirow{2}{*}{Pretrain Dataset} & \multicolumn{3}{c}{RSNA}  \\
    & & 1\% & 10\% & 100\%  \\
    \midrule    
        CLIP \cite{radford2021learning} & ImageNet & 0.348 & 0.399 & 0.640 \\ \addlinespace[1ex]
        ConVIRT \cite{zhang2022contrastive} & MIMIC-CXR & 0.550 & 0.674 & 0.675 \\ \addlinespace[1ex]
        GLoRIA-MIMIC \cite{huang2021gloria} & MIMIC-CXR & 0.603 & \underline{0.687} & 0.683 \\ \addlinespace[1ex]
        LoVT \cite{muller2022joint} & MIMIC-CXR & 0.624 & 0.681 & 0.696  \\ \addlinespace[1ex]
        BioViL \cite{boecking2022making} & MIMIC-CXR & 0.597 & 0.676 & 0.679  \\ \addlinespace[1ex]
        MGCA \cite{wang2022multi} & MIMIC-CXR & \underline{0.630} & 0.683 & \underline{0.698}  \\ \addlinespace[1ex]
    \midrule    
        Ours & MIMIC-CXR & \textbf{0.686} & \textbf{0.696} & \textbf{0.726} \\
    \bottomrule
    \addlinespace[1ex] % 增加空白行，以便表格高度一致
\end{tabular}
}
\caption{Semantic Segmentation results on RSNA Pneumonia and SIIM Pneumothorax datasets with 1\%; 10\%; 100\% training data. \textbf{Bold} denotes the best result and \underline{Underline} denotes the second-best result.}
\label{tab:SemanticSegmentation}
\end{subtable}
\caption{Comparison of results on different tasks.}
\vspace{-3em}
\end{table}

% 4.4.3
\subsubsection{Multi-modal Evaluation}
% 这里少一段描述
Multi-modal evaluation refers to the simultaneous processing and analysis of multiple types of data sources to obtain more comprehensive and accurate information. In this paper, multi-modal evaluation primarily facilitates cross-attention interaction between radiographic images and their associated ``impressions'' through the summarization branch, thereby achieving more precise Visual Question Answering (VQA).

\textbf{Visual Question Answering.} Multi-modal evaluation requires effective interaction between the two modalities. This is particularly faithful in visual question answering (VQA), which aims to generate accurate responses based on visual images and corresponding questions. In our \texttt{HybridMED}, the summarization branch functions as an auxiliary component, facilitating the cross-attention interaction between radiographic images and their associated ``impressions''. Consequently, we retain the visual encoder, the uni-modal text decoder, and the captioning branch to execute VQA. Our concentration is primarily on chest radiographies within both the VQA-RAD and Med-VQA2019 datasets. We adopt the PTUnifier~\cite{chen2023towards} methodology to segregate and train these data, employing two linear layers as the trainable VQA head. As demonstrated in Table \ref{table:VQA}, our \texttt{HybridMED} outperforms other methods in terms of accuracy on both datasets. This confirms the model's ability to concurrently comprehend both visual and textual modalities through the cross-attention mechanism.
%%%%%%%%%%%%%%%%%%%%%%%%%%%%%%%%%%%%%%%%%%%%%%%%%%%%%%%%%%%
% Table-5: VQA 
\begin{table}
\begin{center}
\resizebox{0.85\textwidth}{!}{
%表格中的数据居中，c的个数为表格的列数
\begin{tabular}{cccc} 
	\toprule
 Method & Pretrain Dataset & VQA-RAD-chest & MedVQA-2019-chest \\
 	\midrule	
  
  	MedViLL \cite{moon2022multi} & MIMIC-CXR + OpenI & 0.686 & 0.702 \\
	CPRD \cite{liu2021contrastive} & CRD & 0.683 & 0.678 \\
	MMBERT \cite{khare2021mmbert} & ROCO & 0.672 & 0.696 \\
 	PTUinifier-MIMIC \cite{chen2023towards} & MIMIC-CXR & \underline{0.708} & \underline{0.727} \\
   
  	\midrule	
        Ours & MIMIC-CXR & \textbf{0.747} & \textbf{0.766}
        % \hdashline 
        \\
	\bottomrule
\end{tabular}
}
\end{center}
\caption{VQA results on VQA-RAD-chest and MedVQA-2019-chest datasets. \textbf{Bold} denotes the best result and \underline{Underline} denotes the second-best result.}
\label{table:VQA}
\vspace{-30pt}
\end{table}
% \vspace{-20pt}

%%%%%%%%%%%%%%%%%%%%%%%%%%%%%%%%%%%%%%%%%%%%%%%%%%%%%%%%%%%
%%%%%%%%%%%%%%%%%%%%%%%%%%%%%%%%%%%%%%%%%%%%%%%%%%%%%%%%%%%
\subsection{Qualitative Results}
\begin{figure}[!h]
    \centering
    \begin{minipage}[t]{.4\textwidth}
      \centering
      \includegraphics[width=\linewidth, height=3.5cm]{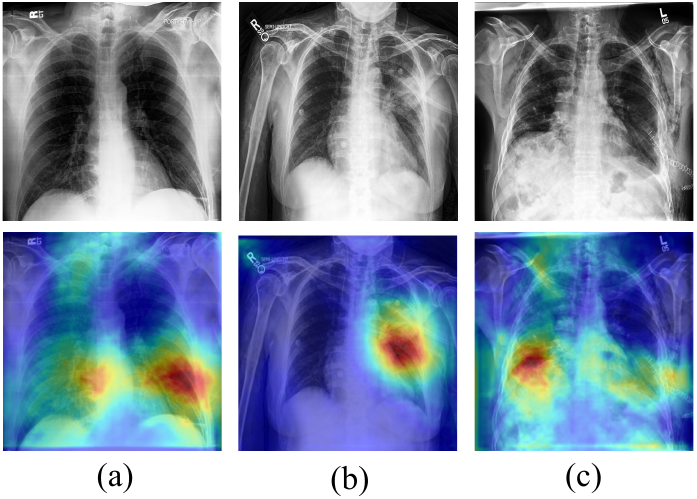}
      \captionof{figure}{Results of cross-modality attention maps visualization. The related prompt is (a) Atelectasis (b) Consolidation and (c) Pleural Effusion.}
      \label{fig:atten}
    \end{minipage}\hfill
    \begin{minipage}[t]{.58\textwidth}
      \centering
      \includegraphics[width=\linewidth, height=3.5cm]{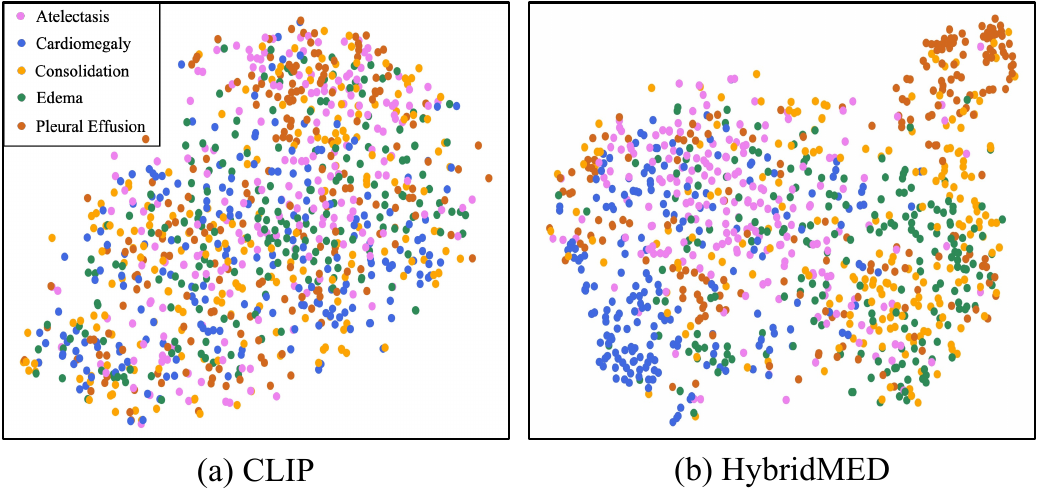}
      \captionof{figure}{t-SNE visualization results on CheXpert 5x200 dataset by CLIP and \texttt{HybridMED}. The figures depict points in various colors, each representing different ground truth disease types along with their corresponding cluster assignments.}
      \label{fig_t_sne}
    \end{minipage}
\end{figure}
% \subsubsection{Visualization}
% As shown in Fig. \ref{fig:atten}, the attention visualization of our HybridMED is presented. Each column represents the same sample, with the first row displaying the original image and the second row showing the model's text-related attention regions. It can be observed that HybridMED accurately focuses on the image regions related to the text, indicating that our model effectively learns the relationships between textual and visual features.

% \subsubsection{t-SNE}
% To further demonstrate the performance, we use t-SNE to visualize the clustering results of CLIP and our HybridMED model for five common chest diseases (Atelectasis, Cardiomegaly, Consolidation, Edema, and Pleural Effusion). As shown in Fig. \ref{fig_t_sne}, compared to CLIP, our HybridMED model better distinguishes these diseases, indicating superior feature representation capability.

To qualitatively assess the performance of our model, we conducted two types of experiments: attention visualization and t-SNE analysis. These experiments demonstrate how effectively our \texttt{HybridMED} model learns and represents the relationships between textual and visual features. As shown in Fig. \ref{fig:atten}, the attention visualization of our \texttt{HybridMED} model is presented. Each column represents the same sample, with the first row displaying the original image and the second row showing the model's text-related attention regions. It can be observed that \texttt{HybridMED} accurately focuses on the image regions related to the text, indicating that our model effectively learns the relationships between textual and visual features. To further demonstrate the model's performance, we employed t-SNE to visualize the clustering results of CLIP and our \texttt{HybridMED} model for five common chest diseases (Atelectasis, Cardiomegaly, Consolidation, Edema, and Pleural Effusion). As shown in Fig. \ref{fig_t_sne}, compared to CLIP, our \texttt{HybridMED} model better distinguishes these diseases, indicating superior feature representation capability.

% As we can see, each part 都发挥了重要的作用
% we use VQA 下游任务

\subsection{Ablation Study}
To verify the effectiveness of different components of our methods, we conducted ablation studies on various parts, as shown in Table \ref{tab:ablation_study}. We evaluated the different components of the \texttt{HybridMED} framework, specifically the impact of contrastive learning, caption generation, summary generation, and knowledge distillation. The experiments were conducted on the RSNA dataset (100\% fine-tuned) and the VQA-RAD dataset, with evaluation metrics including classification, detection, segmentation, zero-shot classification, and visual question answering. When only using contrastive learning or caption generation independently, the performance in classification and zero-shot classification is slightly lower. This is because caption generation is not directly related to pixel-level tasks, potentially introducing additional errors into segmentation. When combining contrastive learning with caption generation, performance improves, suggesting that the integration of these two components has a positive impact on most tasks. However, the inclusion of the summary generation branch did not lead to significant improvements in VQA and even decreased performance in zero-shot classification. This indicates that directly incorporating the summary branch may not effectively serve as a bridging component. The introduction of knowledge distillation validated the effectiveness of the summary branch, demonstrating the necessity of explicit information fusion. The results showed that knowledge distillation brought all tasks to their best performance levels, confirming the overall enhancement in model performance. This validates the effectiveness of multi-modal alignment and generative distillation components. In summary, the results of the ablation study clearly illustrate the roles and importance of each component in the \texttt{HybridMED} framework. While the direct inclusion of the summary branch requires careful handling to avoid performance degradation, the integration of knowledge distillation ensures effective information fusion and enhances the overall model performance across different tasks.

\begin{table*}[!h]
    \begin{center}
   % \resizebox{0.7\linewidth}{!}{
    \begin{tabular}{c c c c | c c c c | c }
    \toprule
    \multicolumn{4}{c}{Training tasks} & \multicolumn{4}{c}{RSNA (100\% fine-tuned)} & \multicolumn{1}{c}{VQA-RAD}\\
    Contrast & Cap & Sum & KD & Cls & Det & Seg & ZS-Cls & VQA \\
    \midrule
    \checkmark & & & & 0.892 & 0.249 & 0.724 & 0.758 &0.736 \\
    & \checkmark & & &  0.893 &  0.245 & 0.702 & 0.739 & 0.735 \\
    \checkmark & \checkmark & & & 0.899 & 0.251 & 0.715 & 0.776 & 0.736 \\
    \checkmark &  \checkmark & \checkmark & &  0.898 & 0.252 & 0.717 & 0.679 & 0.742   \\
    \checkmark & \checkmark & \checkmark & \checkmark & \textbf{0.900}&\textbf{0.256}&\textbf{0.726}&  \textbf{0.800}&\textbf{0.747}
    \\
    \bottomrule
    \end{tabular}
     \end{center}
    \caption{Comparison results of different training tasks on the RSNA (100\% fine-tuned) and VQA-RAD datasets. The training tasks include Contrastive Learning (Contrast), Captioning (Cap), Summarization (Sum), and Knowledge Distillation (KD). The evaluation metrics encompass Classification (Cls), Detection (Det), Segmentation (Seg), Zero-Shot Classification (ZS-Cls), and Visual Question Answering (VQA). \textbf{Bold} denotes the best result.}
    \label{tab:ablation_study}
    \vspace{-30pt}
\end{table*}

\section{Conclusion}
This study proposes \texttt{HybridMED}, a multi-modal contrastive learning pretraining framework for medical image representation learning. By focusing on the hierarchical relationship between ``findings'' and ``impression'' in radiology image datasets, our method effectively aligns global visual representations with ``impression'' and token-level features with ``Findings''. Additionally, we introduce a generative decoder, comprising a description branch and a summary branch, to facilitate knowledge distillation, thereby enhancing the performance of the description branch without significantly increasing parameter complexity. Experimental results across multiple datasets demonstrate that the \texttt{HybridMED} framework achieves substantial performance improvements in various downstream tasks, including classification, segmentation, object detection, and visual question answering tasks. \texttt{HybridMED} showcases the potential of integrating contrastive learning and generative pretraining methods in the medical imaging domain, validated by its superior performance in achieving state-of-the-art results. Comprehensive evaluation of \texttt{HybridMED}, along with qualitative visualizations and t-SNE analysis, highlights its robust feature representation capability, further confirmed by ablation studies on the effectiveness of multi-modal alignment and generative distillation components. Overall, \texttt{HybridMED} marks a significant advancement in Med-VLP methods, offering a versatile and efficient approach to enhance radiological image representation learning and contributing to improved diagnostic processes in medical imaging.

\bibliographystyle{splncs04}
\bibliography{main}
\end{document}